\documentclass[runningheads]{llncs}
\usepackage[T1]{fontenc}
\usepackage{graphicx}
\begin{document}
\title{Civil Court Simulation with Large Language Models}
\author{
Yifan Chen\inst{1} \and
Haitao Li\inst{2} \and
Kaiyuan Zhang\inst{2} \and
Yueyue Wu\inst{2} \and
Qingyao Ai\inst{2}\thanks{Corresponding author.} \and
Yiqun Liu\inst{2}
}

\authorrunning{Y. Chen et al.}
\institute{
Beijing University of Posts and Telecommunications, Beijing, China\\
\email{chenyifan@bupt.edu.cn}
\and
Tsinghua University, Beijing, China\\
\email{aiqy@tsinghua.edu.cn}
}
\maketitle              % typeset the header of the contribution
\begin{abstract}
Court simulation bridges legal education and judicial practice, yet human-based simulations are costly and difficult to scale. Large language models (LLMs) offer a scalable alternative, but existing court-simulation research mainly focuses on criminal cases. Civil litigation is more common in practice and harder to simulate because its claims, liability, and remedies are more flexible. We present a \textbf{multi-agent court simulation framework} for Chinese civil cases. The framework organizes role-based interaction through a five-stage civil trial procedure and integrates memory module and statute retrieval to support long-process adjudication. Experiments show that the framework produces reliable civil judgments, with clear strengths in liability allocation and multi-item adjudication. Further experiments show that memory quality substantially affects downstream simulation quality. Through a five-layer factor framework, we analyze how legal grounding, information conditions, judicial capability and role orientation, organizational pressure, and social context affect the framework's reliability and behavior. These results support the effectiveness of the proposed framework for civil court simulation. The dataset and code are available at: \url{https://github.com/foggpoy/Civil-Court}.

\end{abstract}

\section{Introduction}

Court simulation is an important bridge between legal education and judicial practice. By recreating litigation in a structured setting, it helps reveal how claims are presented, how evidence is examined, and how judgments are formed. Traditional human-based simulations, however, are costly, time-consuming, and difficult to scale. Recent advances in large language models (LLMs) make it increasingly feasible to simulate complex legal interaction with language-based agents, given their strong capabilities in instruction following, role-conditioned interaction, and long-form text generation \cite{lai2023lawsurvey,park2023generative}.

Despite the potential of LLM-based simulations, it remains unclear whether they can adequately capture the dynamics of civil litigation. Existing work has primarily focused on criminal-case simulations, where offense structures, evidentiary standards, and factual disputes are relatively well defined, making them suitable early testbeds for LLM-based court simulation ~\cite{zhang2025simcourt}. In contrast, civil litigation, which is more common in judicial practice, involves more flexible evidentiary requirements, greater uncertainty in argumentation, and more heterogeneous liability and remedy structures. These characteristics make civil trials harder to simulate and a valuable domain for studying courtroom dynamics and judicial decision processes with LLM-based simulation. They also make direct fact-based judgment generation difficult, motivating a staged multi-agent design that organizes claims, evidence, and disputed issues before adjudication.

Recent work has shown the promise of LLMs in legal generation, retrieval, and agent-based interaction \cite{lai2023lawsurvey,li2024lecardv2,luo2025largelanguagemodelagent}. Some studies have also begun to explore legal process simulation with LLM agents \cite{zhang2025simcourt,chen2024agentcourt,chen2025agentmediation}. However, civil adjudication remains relatively underexplored, despite its greater variability in claims, liability allocation, and dispositive structure.

We present a framework for \emph{civil court simulation with large language models}. The framework models three courtroom roles---judge, plaintiff, and defendant---and organizes their interaction around a five-stage civil trial procedure. It further incorporates a memory module and statute retrieval to support long-process adjudication and final judgment generation. To support the experiments, we also adapt real Chinese civil case materials from an existing benchmark into a simulation-oriented dataset ~\cite{li2025casegen}.

We evaluate the framework from both effectiveness and analytical perspectives. Because the final judgment is the most important output of the simulated procedure, the evaluation centers on judgment quality under an LLM-as-a-Judge setting ~\cite{gu2026llmjudge}. The results show that the proposed framework produces more reliable civil judgments than direct fact-based generation, with clear gains in liability allocation and multi-item judgment organization. We further find that memory quality substantially affects downstream simulation quality. Finally, we use a five-layer factor framework to analyze how legal grounding, information conditions, role orientation, and external pressure influence the framework's behavior. The systematic and interpretable changes observed under these controlled interventions further support its reliability as an experimental environment for studying civil adjudication.

Our main contributions are as follows:
\begin{itemize}
    \item We present a multi-agent framework for civil court simulation with LLMs, centered on role-based interaction across a five-stage civil trial procedure.
    \item We verify the effectiveness of the proposed framework on Chinese civil cases and show that memory quality is a key factor in downstream simulation quality.
    \item We propose a five-layer factor framework for analyzing simulated adjudication and use it to study how legal, informational, individual, organizational, and social factors affect the civil court simulation framework.
\end{itemize}

\section{Related Work}

Existing research provides the technical and methodological foundation for LLM-based civil court simulation. Advances in legal LLMs, retrieval-augmented legal generation, and LLM-based agents show that language models can support legal reasoning, external knowledge grounding, and sustained role-based interaction. Recent work on legal simulation and LLM-based evaluation further suggests that LLMs can be used not only to generate legal texts, but also to model legal processes and assess complex legal outputs.

\paragraph{Legal LLMs and Agents.}
Recent research on legal large language models has expanded legal NLP from task-specific prediction to broader legal reasoning, generation, and assistance ~\cite{lai2023lawsurvey}. In parallel, LLM-based agent research has shown that language models can support role-conditioned interaction, memory, planning, and multi-agent collaboration \cite{park2023generative,wu2023autogenenablingnextgenllm}. These developments make it possible to model courtroom procedures as sustained interaction among legal roles rather than isolated text generation.

\paragraph{Legal Retrieval.}
Legal reasoning often requires grounding in external legal materials rather than relying only on the parametric knowledge of LLMs. Retrieval-augmented generation connects language models with external knowledge sources and has become a common approach for knowledge-intensive generation ~\cite{lewis2020retrieval}. In the Chinese legal domain, LeCaRDv2 advances legal case retrieval, while LexRAG studies statute retrieval in multi-turn legal consultation and shows that retrieved legal articles can support downstream legal generation \cite{li2024lecardv2,li2025lexrag}. Our framework follows this line by using statute retrieval as a grounding component in the simulated trial process.

\paragraph{Legal Simulation.}
Recent work has begun to use LLM agents for legal process simulation. SimCourt proposes a Chinese criminal court simulation framework ~\cite{zhang2025simcourt}. AgentCourt studies courtroom interaction through adversarial evolution of lawyer agents across repeated simulated cases ~\cite{chen2024agentcourt}. AgentMediation extends LLM-based legal simulation to dispute mediation and uses controlled experiments to analyze legal and social factors ~\cite{chen2025agentmediation}. These studies demonstrate the potential of LLM agents for legal simulation. However, existing work has mainly focused on criminal proceedings, lawyer-agent training, mediation, and related settings. Civil court simulation remains less explored, although civil litigation is more flexible in evidentiary requirements, more uncertain in argumentation, and more diverse in claims, remedies, and liability structures.

\paragraph{LLM-as-a-Judge.}
Evaluating civil court simulation is challenging because civil judgments are open-ended, case-dependent, and structurally heterogeneous. LLM-as-a-Judge provides a practical way to assess complex outputs with explicit evaluation criteria \cite{gu2026llmjudge,li2025generationjudgment}. This supports our use of a structured LLM-based evaluation design for simulated civil judgments, where simple lexical metrics or fixed-field matching are insufficient.

\section{Method}

Our proposed method simulates civil adjudication as a role-based and staged courtroom process. It consists of four components: data construction, a multi-agent civil court simulation framework, judgment evaluation, and a five-layer factor framework for controlled analysis.

\subsection{Data Construction}

The data construction aims to convert existing civil case materials into procedural inputs for court simulation. We build the experimental data based on CaseGen, which provides real Chinese legal case materials for multi-stage legal case document generation~\cite{li2025casegen}. Since the original format is not designed for courtroom simulation, we reorganize the materials to support role initialization, claim and defense presentation, evidence examination, debate, and final judgment generation.

Each case instance contains party information, complaint, statement of defense, case facts, evidence, and a reference judgment. The reference judgment is not provided to any agent during simulation and is used only for evaluation. We select 100 civil cases for the main experiments.

\subsection{Civil Court Simulation Framework}

\begin{figure}
\includegraphics[width=\textwidth]{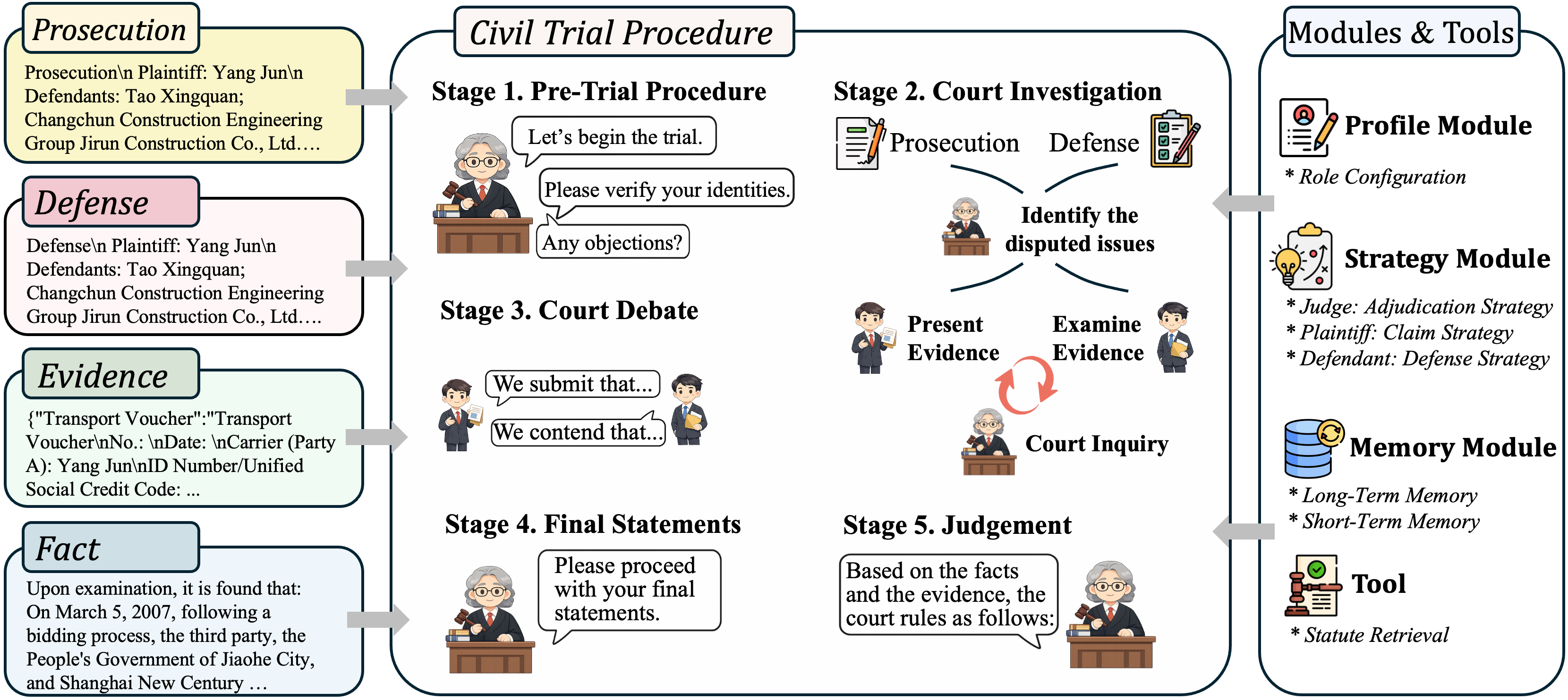}
\caption{Overview of the multi-agent civil court simulation framework.}
\label{fig:overview}
\end{figure}

The civil court simulation framework models adjudication as a five-stage interaction process. As shown in Figure~\ref{fig:overview}, the framework takes a case instance as input, initializes the courtroom roles, runs the trial procedure stage by stage, maintains memory, retrieves relevant statutes when needed, and finally generates a structured civil judgment.

\subsubsection{Role Configuration}

The framework contains three core agents: judge, plaintiff, and defendant. Each agent is defined by a role configuration and a strategy specification. The role configuration determines the procedural identity of the agent, while the strategy specification guides how the agent participates in the simulated trial.

The judge acts as the presiding adjudicator. This role maintains trial order, identifies disputed issues, examines evidence, organizes debate, and generates the final judgment. Its strategy focuses on fact-finding, responsibility determination, evidence review, compensation assessment, and unresolved issues.

The plaintiff acts as the party initiating the civil action. This role presents claims, explains factual grounds, supports requested remedies, and responds to the defendant's challenges. The defendant acts as the responding party. This role contests the plaintiff's evidence, challenges liability, presents favorable facts, and seeks to reduce or avoid responsibility. Together, these roles create an adversarial setting in which the final judgment is shaped by both party-side interaction and judicial organization.

\subsubsection{Trial Procedure}

The simulation follows the structure of Chinese civil trial procedure and is organized into five stages: pre-trial procedure, court investigation, court debate, final statements, and judgment. This design allows the framework to preserve the main procedure of real civil hearings while keeping the process controllable for systematic experiments.

The pre-trial procedure stage establishes the procedural starting point. The judge opens the hearing, verifies the identities of the parties, and asks routine procedural questions such as whether recusal is requested.

The court investigation stage develops the factual basis of the case. The plaintiff states the claims and factual grounds, and the defendant provides a response. The judge then identifies disputed issues. The process continues with evidence presentation and examination: the plaintiff presents evidence and explains what each item is intended to prove, while the defendant examines the evidence and raises objections when necessary. The judge may also conduct additional inquiry when key factual issues remain unclear.

The court debate stage focuses on adversarial argumentation. The plaintiff and defendant argue over liability, legal application, remedies, and the interpretation of evidence. When unresolved issues remain, the judge may organize additional focused debate. This stage helps the framework model the argumentative dynamics of civil litigation.

The final statements stage allows both parties to make concluding remarks before judgment. In the judgment stage, the judge generates a structured civil judgment based on the case materials, the accumulated trial information, and retrieved statutes when retrieval is used.

\subsubsection{Memory Module}

The memory module preserves judgment-relevant information across the simulated trial. Civil court simulation is a long-process task, and directly passing the full raw transcript to later stages may introduce redundancy and noise. We therefore use both short-term and long-term memory.

\textbf{\textit{Short-term memory}} refers to the dialogue history within the current stage. It supports local coherence when an agent responds to an immediately preceding statement, objection, or question. \textbf{\textit{Long-term memory}} refers to structured summaries generated after important stages, especially court investigation and court debate. The memory preserves disputed issues, evidence-related findings, party disagreements, responsibility relations, and key numerical information. Later stages, especially debate and judgment, use it to maintain procedural continuity.

\subsubsection{Statute Retrieval}

The framework includes a statute retrieval component for legal grounding. The retrieval tool is backed by a Chinese statute library containing more than 17,000 commonly used legal provisions. Each statute is embedded with \texttt{text-embedding-v4}, indexed with FAISS, and retrieved through nearest-neighbor search.

Retrieval is used as an auxiliary component rather than as a replacement for judicial reasoning. Before retrieval, the corresponding role determines whether external legal support is needed and formulates a natural-language query. In the debate stage, retrieval can support party-side legal argumentation. In the judgment stage, the judge can retrieve statutes based on the case materials and accumulated trial information. Retrieved statutes are appended to the generation context as external legal support.

\subsection{Evaluation}

The final judgment is used as the primary evaluation target because it integrates claims, disputed issues, evidence examination, court debate, and legal application. Errors in earlier stages are expected to be reflected in the final judgment, making judgment quality a compact indicator of the overall simulation process.

We adopt an LLM-as-a-Judge evaluation design because civil judgments are open-ended and structurally heterogeneous. Unlike criminal simulation, where outcomes can often be mapped to relatively stable targets such as imprisonment terms or fines, civil judgments may involve different claims, liable parties, forms of liability, monetary outcomes, and multiple dispositive items. Simple lexical metrics or fixed-field matching are therefore insufficient.

The evaluator compares each generated judgment with the corresponding real judgment and scores it on five dimensions: Judgment Conclusion Consistency (JCC), Accuracy of Liable Parties and Liability Allocation (ALA), Quantitative Judgment Precision (QJP), Appropriateness of Legal Basis (ALB), and Accuracy in Multi-Party/Multi-Item Adjudication (AMA). The overall score is computed as:
\begin{equation}
S = 0.35s_{\mathrm{JCC}} + 0.20s_{\mathrm{ALA}} + 0.25s_{\mathrm{QJP}} + 0.10s_{\mathrm{ALB}} + 0.10s_{\mathrm{AMA}}
\end{equation}
The weights emphasize dispositive correctness, quantitative precision, and liability allocation, which are central to civil judgment quality. Legal basis and multi-item handling are assigned smaller but necessary weights to capture legal grounding and structural completeness. We use Kimi-K2.5~\cite{moonshot2026kimik25} as the evaluator model. The evaluator is run with temperature 0.0 to reduce generation variance.

\subsection{Five-Layer Factor Framework}

\begin{figure}
\includegraphics[width=\textwidth]{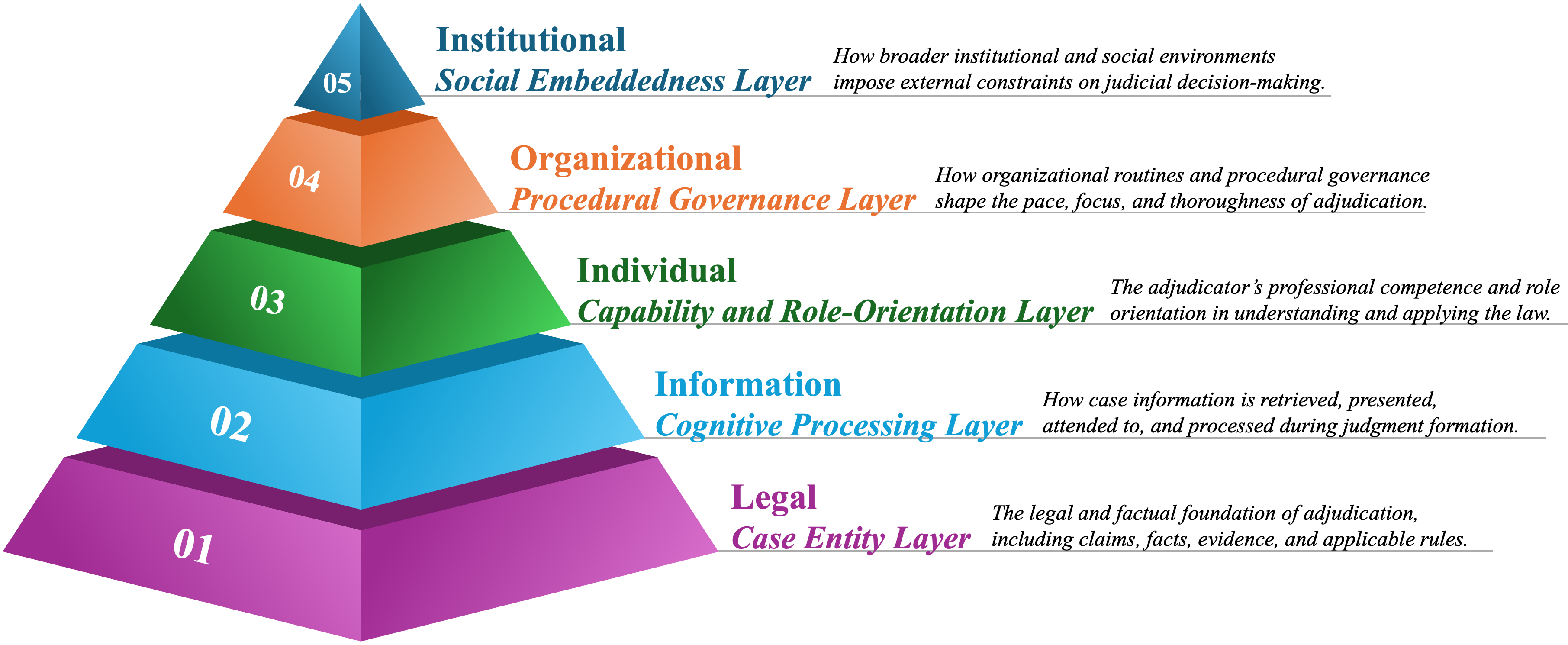}
\caption{Five-layer factor framework for controlled analysis of civil court simulation.}
\label{fig:pyramid}
\end{figure}

The framework is also used to analyze how different conditions affect civil court simulation. Since the simulation pipeline is modular, specific conditions can be introduced or removed while keeping the underlying case materials and trial procedure unchanged. To organize these interventions, we construct a five-layer factor framework, shown in Figure~\ref{fig:pyramid}. The framework follows a bottom-up logic: lower layers are closer to the normative basis of adjudication, such as facts, evidence, and law, while upper layers correspond to broader organizational and social conditions. This design allows us to examine how legal grounding, information conditions, role settings, organizational pressure, and social context influence the behavior and reliability of the simulation framework.

\textbf{Legal-Case Entity Layer} concerns the direct legal and factual basis of adjudication, including facts, evidence, statutes, judicial interpretations, and other legal materials. Interventions at this layer examine whether weakening legal or factual grounding changes simulation quality, such as removing statute retrieval or removing court-established facts from the judgment input.

\textbf{Information-Cognitive Processing Layer} concerns what information the adjudicator can access and how that information is presented. In our framework, this includes retrieval coverage, evidence order, information salience, and party-side presentation strength. Interventions include restricting the statute retrieval library, strengthening one party's role configuration, and changing the order of evidence presentation.

\textbf{Individual-Capability and Role-Orientation Layer} concerns model capability and judicial role orientation. We operationalize it through model replacement and modifications to the judge's role configuration. These settings test whether the simulation is sensitive to role capability and judicial orientations such as formalism, substantivism, empathy, and impatience.

\textbf{Organizational-Procedural Governance Layer} concerns organizational and procedural pressures beyond case merits, such as efficiency goals, case-closing requirements, workflow constraints, and internal management expectations. In this work, we examine this layer through a performance-oriented judge configuration that emphasizes case-closing efficiency.

\textbf{Institutional-Social Embeddedness Layer} concerns broader institutional and social conditions, such as public opinion, media attention, and perceived social acceptability. We examine this layer by introducing public-opinion pressure into the judge's role configuration.

\section{Experiments and Analysis}

The experiments examine three aspects of the proposed civil court simulation framework. First, we evaluate whether the framework can produce reliable civil judgments. Second, we analyze the role of the memory module in the framework. Third, we use the five-layer factor framework to study how different conditions affect the reliability and behavior of the civil court simulation framework.

All numerical results are reported as mean scores. The main results are evaluated on the 100-case test set, while the remaining analysis experiments are conducted on a 30-case subset. We use the following abbreviations: JCC = Judgment Conclusion Consistency, ALA = Accuracy of Liable Parties and Liability Allocation, QJP = Quantitative Judgment Precision, ALB = Appropriateness of Legal Basis, AMA = Accuracy in Multi-Party/Multi-Item Adjudication, and Tot. = weighted total score.

\subsection{Main Results}

\begin{figure}
\centering
\includegraphics[width=\textwidth]{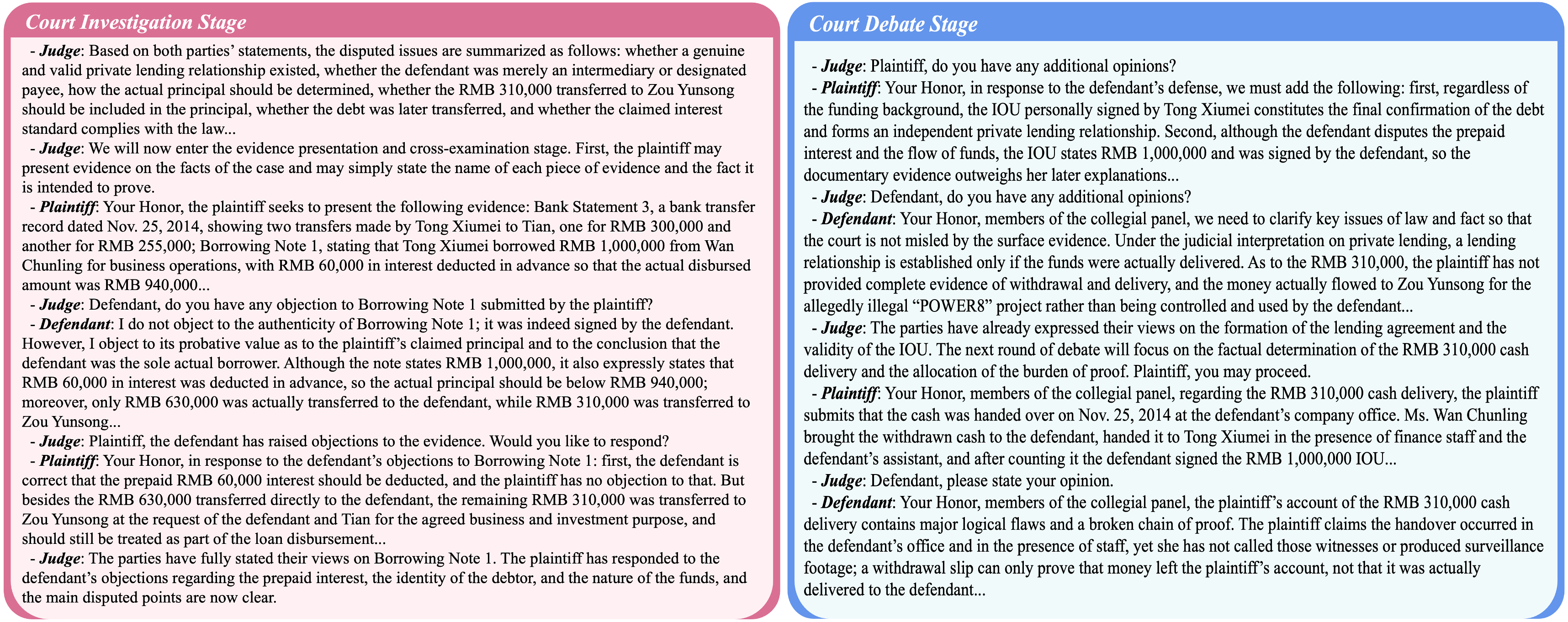}
\caption{Example of role-based interaction in the simulated civil trial.}
\label{fig:example}
\end{figure}

Figure~\ref{fig:example} provides an example of role-based interaction in the simulated civil trial process. Table~\ref{tab:effectiveness} further reports whether such staged interaction improves final judgment quality.

In the Kimi setting, the fact-only baseline directly generates the final judgment from case facts using Kimi-K2.5 ~\cite{moonshot2026kimik25}. The simulation setting uses Kimi-K2.5 as the judge, Qwen3.5-27B as the plaintiff and defendant, and Kimi-K2.5 as the summary model \cite{moonshot2026kimik25,qwen2026qwen35}. Therefore, the main difference is whether the staged civil trial process is used.

\begin{table}
\caption{Effectiveness comparison between direct generation and civil court simulation.}
\label{tab:effectiveness}
\centering
\small
\setlength{\tabcolsep}{4pt}
\begin{tabular}{lrrrrrr}
\hline
Setting & JCC & ALA & QJP & ALB & AMA & Tot. \\
\hline
Kimi-Fact & 5.31 & 5.44 & 4.45 & 5.54 & 4.98 & 5.11 \\
Kimi-Sim  & 5.25 & 5.86 & \textbf{4.47} & \textbf{5.60} & 5.05 & \textbf{5.19} \\
\hline
DS-Fact   & 4.86 & 5.26 & 3.77 & 5.12 & 4.78 & 4.68 \\
DS-Sim    & 5.15 & 5.71 & 4.21 & 5.25 & 5.06 & \textbf{5.03} \\
GLM-Fact  & 5.15 & 5.55 & 4.15 & 5.34 & 4.94 & 4.98 \\
GLM-Sim   & \textbf{5.46} & \textbf{5.96} & 4.45 & 5.59 & \textbf{5.34} & \textbf{5.31} \\
\hline
\end{tabular}
\end{table}

The simulation framework improves the total score from 5.11 to 5.19 in the main Kimi setting. The improvement is modest but concentrated in ALA, which rises from 5.44 to 5.86. QJP, ALB, and AMA also show small gains, while JCC slightly decreases. This suggests that staged simulation is especially helpful for responsibility mapping, but amount-sensitive reasoning and exact conclusion consistency remain challenging.

The cross-model results show a stronger tendency. Under the DeepSeek configuration, the total score increases from 4.68 to 5.03~\cite{deepseek2025v32}. Under the GLM configuration, the total score increases from 4.98 to 5.31, with improvements across all five dimensions~\cite{zeng2026glm5}. These results suggest that the framework provides a structural advantage across different judge models, although the magnitude of improvement depends on how effectively the judge model uses information accumulated during investigation, evidence examination, and debate.

Overall, the results show that civil judgment generation remains challenging for LLMs, especially in direct fact-to-judgment generation. By decomposing the process into role-based trial stages, the framework helps the judge model organize claims, defenses, evidence, and disputed issues before judgment generation. The clearest gains appear in liability allocation and multi-item adjudication, where civil judgments are particularly complex.

\subsection{Role of the Memory Module}

The next experiment evaluates the role of long-term memory. The memory mechanism is kept unchanged, while the summary model used to generate long-term memory after key stages is replaced. Table~\ref{tab:memory} reports the results.

\begin{table}
\caption{Effect of the summary model.}
\label{tab:memory}
\centering
\small
\setlength{\tabcolsep}{4pt}
\begin{tabular}{lrrrrrr}
\hline
Model & JCC & ALA & QJP & ALB & AMA & Tot. \\
\hline
Kimi-K2.5 & \textbf{5.53} & \textbf{6.17} & \textbf{4.73} & \textbf{5.77} & \textbf{5.43} & \textbf{5.47} \\
Qwen3-32B & 5.10 & 5.70 & 4.27 & 5.50 & 5.27 & 5.07 \\
Qwen3-4B  & 5.13 & 5.43 & 4.23 & 5.47 & 5.10 & 5.00 \\
\hline
\end{tabular}
\end{table}

Replacing only the summary model causes a clear decline in final judgment quality. The total score drops by 0.40 with the Qwen3-32B summary model and by 0.47 with the Qwen3-4B summary model ~\cite{yang2025qwen3}. The largest declines occur in ALA, QJP, and JCC, while ALB changes less.

This result shows that the memory module is not merely a context-compression tool. Long-term memory carries disputed issues, responsibility relations, evidence-related findings, and numerical conditions into later stages. When memory quality declines, the framework becomes less stable in responsibility mapping, dispositive organization, and numerical determination. The effectiveness of staged simulation therefore depends not only on having multiple trial stages, but also on preserving judgment-relevant structure across those stages.

\subsection{Factor Analysis}

The five-layer factor framework is used to analyze how different conditions affect the simulation framework. Unless otherwise specified, the baseline configuration uses Kimi-K2.5 as the judge, Qwen3.5-27B as the plaintiff and defendant, and Kimi-K2.5 as the summary model \cite{moonshot2026kimik25,qwen2026qwen35}. Table~\ref{tab:factor} reports the main controlled intervention results.

% \noindent
% \textit{Notes.} NR = no statute retrieval; NF = no court-established facts in final judgment input; LR = local regulation emphasis; CC = retrieval restricted to the Civil Code; PS = stronger plaintiff; DS = stronger defendant; EO = changed evidence order; F = formalist judge; S = substantive judge; E = high-empathy judge; I = impatient judge; PO = performance orientation; OP = public-opinion pressure.

The Legal-Case Entity Layer shows that legal and factual grounding are important anchors of simulated adjudication. Removing statute retrieval reduces the total score from 5.47 to 5.09, while removing court-established facts reduces it to 4.95. These declines indicate that when the relevant legal materials and established facts available to the judge are weakened, the simulated judgment becomes less stable. This result is consistent with the legal model's emphasis on judicial decision-making through the interpretation of relevant legal materials ~\cite{Brenner_Whitmeyer_2009}. At the same time, the framework does not collapse without retrieval, suggesting that the model still retains some adjudicative capacity from case facts and prior legal knowledge. Emphasizing local regulations produces little overall change, indicating that local norms play a more case-contingent role in the present dataset. Across these interventions, QJP remains low, which is consistent with prior work showing instability in judicial numerical judgments ~\cite{Rachlinski2015NumericJudgments}.

\begin{table}
\caption{Controlled factor intervention results under the baseline configuration.}
\label{tab:factor}
\centering
\footnotesize
\setlength{\tabcolsep}{4pt}
\begin{tabular}{llrrrrrr}
\hline
Layer & Setting & JCC & ALA & QJP & ALB & AMA & Tot. \\
\hline
Base & Base & \textbf{5.53} & \textbf{6.17} & \textbf{4.73} & \textbf{5.77} & \textbf{5.43} & \textbf{5.47} \\
\hline
Legal & NR & 5.17 & 5.50 & 4.43 & 5.57 & 5.17 & 5.09 \\
Legal & NF & 5.00 & 5.50 & 4.33 & 5.27 & 4.93 & 4.95 \\
Legal & LR & 5.60 & 6.07 & 4.63 & 5.67 & 5.43 & 5.44 \\
\hline
Information & CC & 5.43 & 6.13 & 4.33 & 5.73 & 5.27 & 5.31 \\
Information & PS & \underline{5.97} & \underline{6.63} & \underline{4.90} & \underline{5.80} & \underline{6.00} & \underline{5.82} \\
Information & DS & 4.90 & 5.50 & 4.27 & 5.47 & 4.83 & 4.91 \\
Information & EO & 5.23 & 5.63 & 4.20 & 5.57 & 5.27 & 5.09 \\
\hline
Individual & F & 4.93 & 5.57 & 3.87 & 5.33 & 5.00 & 4.84 \\
Individual & S & 4.60 & 5.03 & 3.93 & 5.03 & 4.63 & 4.57 \\
Individual & E & 4.97 & 5.47 & 4.27 & 5.20 & 4.90 & 4.91 \\
Individual & I & 4.67 & 5.40 & 3.77 & 5.30 & 4.73 & 4.66 \\
\hline
Organization & PO & 4.90 & 5.60 & 4.07 & 5.23 & 5.13 & 4.89 \\
Institution & OP & 5.27 & 5.53 & 4.50 & 5.50 & 5.17 & 5.14 \\
\hline
\end{tabular}

\vspace{3pt}
\begin{minipage}{0.98\textwidth}
\scriptsize
\textit{Notes.} NR = no statute retrieval; NF = no court-established facts; LR = local regulation emphasis; CC = Civil Code retrieval only; PS = stronger plaintiff; DS = stronger defendant; EO = changed evidence order; F = formalist judge; S = substantive judge; E = high-empathy judge; I = impatient judge; PO = performance orientation; OP = public-opinion pressure.
\end{minipage}
\end{table}

The Information-Cognitive Processing Layer shows that judgment quality is sensitive to how information is supplied and organized. Restricting retrieval to the Civil Code moderately reduces the total score to 5.31, suggesting that the Civil Code often supports the general liability framework but may be insufficient for fine-grained monetary or multi-item decisions. Changing evidence order lowers the total score to 5.09. This is consistent with research showing that order effects are common in legal decision-making ~\cite{WojciechowskiWhiteAllefeldPothos2025OrderEffects}. The stronger-plaintiff condition improves the total score to 5.82, while the stronger-defendant condition reduces it to 4.91. This asymmetry does not imply that plaintiff-side arguments are inherently more correct. Rather, it reflects the procedural role of the plaintiff in civil litigation: clearer plaintiff-side articulation can make claims, requested remedies, and responsibility structures more explicit. This interpretation is also compatible with empirical work showing that plaintiff demands can function as anchoring signals in adjudication ~\cite{ChangChenLiaoLin2023AskMoreAwardedMore}.

The Individual-Capability and Role-Orientation Layer shows that the framework is sensitive to both judicial capability and role orientation. The judge-model replacement experiments in Table~\ref{tab:judge_model} show that Qwen3.5-397B-A17B and GLM-5 both reach a total score of 5.39, while Qwen3-4B reaches only 4.54 \cite{qwen2026qwen35,zeng2026glm5,yang2025qwen3}. These differences indicate that different judge models have different capacities to handle complex civil adjudication, especially in integrating disputed issues, legal rules, evidence, and remedies into a stable judgment structure. The result is consistent with bounded rationality research, which argues that judges use rational means to pursue goals but remain constrained by their own cognitive capacity and institutional environment ~\cite{Fay2024OutOfBounds}. Role-orientation interventions further show that formalist, substantive, high-empathy, and impatient judge configurations all reduce performance. The judge's role orientation therefore affects not only expression style, but also the substantive organization of the final ruling.

\begin{table}
\caption{Judge model replacement results with the plaintiff, defendant, and summary models fixed to Qwen3-32B.}
\label{tab:judge_model}
\centering
\footnotesize
\setlength{\tabcolsep}{4pt}
\begin{tabular}{lrrrrrr}
\hline
Judge Model & JCC & ALA & QJP & ALB & AMA & Tot. \\
\hline
Kimi-K2.5 & 4.83 & 5.47 & 4.23 & 5.27 & 5.07 & 4.88 \\
DeepSeek-V3.2 & 4.83 & 5.23 & 4.20 & 5.13 & 4.73 & 4.78 \\
Qwen3.5-397B-A17B & 5.57 & \textbf{6.03} & 4.53 & 5.27 & \textbf{5.70} & \textbf{5.39} \\
Qwen3-Max & 4.77 & 5.27 & 4.23 & 5.17 & 4.77 & 4.77 \\
Qwen3-4B & 4.77 & 5.00 & 3.90 & 4.77 & 4.17 & 4.54 \\
GLM-5 & \textbf{5.67} & 5.90 & \textbf{4.57} & \textbf{5.57} & 5.23 & \textbf{5.39} \\
\hline
\end{tabular}
\end{table}

The upper two layers show that external constraints also affect the civil trial framework. Performance orientation reduces the total score to 4.89, with especially clear declines in QJP and JCC. This suggests that efficiency-oriented governance can compress the deliberative process and weaken careful numerical and dispositive organization, which is consistent with research on judicial managerialization ~\cite{ColauxSchiffinoMoyson2023Managerialization}. Public-opinion pressure reduces the total score to 5.14. The decline is more moderate but distributed across dimensions, suggesting that public opinion introduces an additional concern for social acceptability rather than replacing legal reasoning outright. This is consistent with research showing that public sentiment can affect judicial behavior and opinion writing \cite{BlackOwensWedekingWohlfarth2016OpinionClarity}.

Taken together, the factor experiments support both the analytical value of the five-layer framework and the reliability of the simulation framework. Most interventions lead to systematic and interpretable changes: legal and factual grounding affects the normative basis of judgment; information conditions affect how the case is represented to the adjudicator; model capability and role orientation affect issue integration; organizational and social pressure shifts adjudicative attention toward efficiency or external acceptability. These results show that the five-layer framework can support controlled analysis of how different conditions affect the reliability and behavior of the civil court simulation framework.

\section{Conclusion}

We present a multi-agent framework for civil court simulation with large language models. The framework models the judge, plaintiff, and defendant as separate courtroom roles, organizes their interaction around a five-stage civil trial procedure, and integrates memory and statute retrieval to support long-process adjudication in the Chinese civil litigation setting. Experiments show that the framework produces reliable civil judgments. The memory experiments further show that memory quality is a key factor in downstream simulation quality. Through a five-layer factor framework, we conduct controlled analysis of how legal grounding, information conditions, judicial capability and role orientation, organizational pressure, and social context affect the reliability and behavior of the simulation framework. The systematic and interpretable results support the effectiveness of the proposed framework for civil court simulation.

\begin{credits}
% \subsubsection{\ackname} A bold run-in heading in small font size at the end of the paper is
% used for general acknowledgments, for example: This study was funded

\subsubsection{\discintname}
% It is now necessary to declare any competing interests or to specifically
% state that the authors have no competing interests. Please place the
% statement with a bold run-in heading in small font size beneath the
% (optional) acknowledgments\footnote{If EquinOCS, our proceedings submission
% system, is used, then the disclaimer can be provided directly in the system.},
% for example: The authors have no competing interests to declare that are
% relevant to the content of this article. Or: Author A has received research
% grants from Company W. Author B has received a speaker honorarium from
% Company X and owns stock in Company Y. Author C is a member of committee Z.
The authors have no competing interests to declare that are
relevant to the content of this article.
\end{credits}
\bibliographystyle{splncs04}
\bibliography{references}

@inproceedings{park2023generative,
author = {Park, Joon Sung and O'Brien, Joseph and Cai, Carrie Jun and Morris, Meredith Ringel and Liang, Percy and Bernstein, Michael S.},
title = {Generative Agents: Interactive Simulacra of Human Behavior},
year = {2023},
isbn = {9798400701320},
publisher = {Association for Computing Machinery},
address = {New York, NY, USA},
url = {https://doi.org/10.1145/3586183.3606763},
doi = {10.1145/3586183.3606763},
booktitle = {Proceedings of the 36th Annual ACM Symposium on User Interface Software and Technology},
articleno = {2},
numpages = {22},
location = {San Francisco, CA, USA},
series = {UIST '23}
}

@misc{luo2025largelanguagemodelagent,
      title={Large Language Model Agent: A Survey on Methodology, Applications and Challenges}, 
      author={Junyu Luo and Weizhi Zhang and Ye Yuan and Yusheng Zhao and Junwei Yang and Yiyang Gu and Bohan Wu and Binqi Chen and Ziyue Qiao and Qingqing Long and Rongcheng Tu and Xiao Luo and Wei Ju and Zhiping Xiao and Yifan Wang and Meng Xiao and Chenwu Liu and Jingyang Yuan and Shichang Zhang and Yiqiao Jin and Fan Zhang and Xian Wu and Hanqing Zhao and Dacheng Tao and Philip S. Yu and Ming Zhang},
      year={2025},
      eprint={2503.21460},
      archivePrefix={arXiv},
      primaryClass={cs.CL},
      url={https://arxiv.org/abs/2503.21460}, 
}

@misc{zhang2025simcourt,
      title={Chinese Court Simulation with LLM-Based Agent System}, 
      author={Kaiyuan Zhang and Jiaqi Li and Yueyue Wu and Haitao Li and Cheng Luo and Shaokun Zou and Yujia Zhou and Weihang Su and Qingyao Ai and Yiqun Liu},
      year={2025},
      eprint={2508.17322},
      archivePrefix={arXiv},
      primaryClass={cs.CY},
      url={https://arxiv.org/abs/2508.17322}, 
}

@misc{li2025casegen,
      title={CaseGen: A Benchmark for Multi-Stage Legal Case Documents Generation}, 
      author={Haitao Li and Jiaying Ye and Yiran Hu and Jia Chen and Qingyao Ai and Yueyue Wu and Junjie Chen and Yifan Chen and Cheng Luo and Quan Zhou and Yiqun Liu},
      year={2025},
      eprint={2502.17943},
      archivePrefix={arXiv},
      primaryClass={cs.CL},
      url={https://arxiv.org/abs/2502.17943}, 
}

@article{gu2026llmjudge,
  title = {A survey on LLM-as-a-judge},
  author = {Jiawei Gu and Xuhui Jiang and Zhichao Shi and Hexiang Tan and Xuehao Zhai and Chengjin Xu and Wei Li and Yinghan Shen and Shengjie Ma and Honghao Liu and Saizhuo Wang and Kun Zhang and Zhouchi Lin and Bowen Zhang and Lionel Ni and Wen Gao and Yuanzhuo Wang and Jian Guo},
  journal = {The Innovation},
  year = {2026},
  pages = {101253},
  issn = {2666-6758},
  doi = {10.1016/j.xinn.2025.101253},
  url = {https://www.sciencedirect.com/science/article/pii/S2666675825004564}
}

@misc{li2025generationjudgment,
      title={From Generation to Judgment: Opportunities and Challenges of LLM-as-a-judge}, 
      author={Dawei Li and Bohan Jiang and Liangjie Huang and Alimohammad Beigi and Chengshuai Zhao and Zhen Tan and Amrita Bhattacharjee and Yuxuan Jiang and Canyu Chen and Tianhao Wu and Kai Shu and Lu Cheng and Huan Liu},
      year={2025},
      eprint={2411.16594},
      archivePrefix={arXiv},
      primaryClass={cs.AI},
      url={https://arxiv.org/abs/2411.16594}, 
}

@misc{wu2023autogenenablingnextgenllm,
      title={AutoGen: Enabling Next-Gen LLM Applications via Multi-Agent Conversation}, 
      author={Qingyun Wu and Gagan Bansal and Jieyu Zhang and Yiran Wu and Beibin Li and Erkang Zhu and Li Jiang and Xiaoyun Zhang and Shaokun Zhang and Jiale Liu and Ahmed Hassan Awadallah and Ryen W White and Doug Burger and Chi Wang},
      year={2023},
      eprint={2308.08155},
      archivePrefix={arXiv},
      primaryClass={cs.AI},
      url={https://arxiv.org/abs/2308.08155}, 
}

@inproceedings{lewis2020retrieval,
author = {Lewis, Patrick and Perez, Ethan and Piktus, Aleksandra and Petroni, Fabio and Karpukhin, Vladimir and Goyal, Naman and K\"{u}ttler, Heinrich and Lewis, Mike and Yih, Wen-tau and Rockt\"{a}schel, Tim and Riedel, Sebastian and Kiela, Douwe},
title = {Retrieval-augmented generation for knowledge-intensive NLP tasks},
year = {2020},
isbn = {9781713829546},
publisher = {Curran Associates Inc.},
address = {Red Hook, NY, USA},
booktitle = {Proceedings of the 34th International Conference on Neural Information Processing Systems},
articleno = {793},
numpages = {16},
location = {Vancouver, BC, Canada},
series = {NIPS '20}
}

@misc{lai2023lawsurvey,
      title={Large Language Models in Law: A Survey}, 
      author={Jinqi Lai and Wensheng Gan and Jiayang Wu and Zhenlian Qi and Philip S. Yu},
      year={2023},
      eprint={2312.03718},
      archivePrefix={arXiv},
      primaryClass={cs.CL},
      url={https://arxiv.org/abs/2312.03718}, 
}

@inproceedings{li2024lecardv2,
author = {Li, Haitao and Shao, Yunqiu and Wu, Yueyue and Ai, Qingyao and Ma, Yixiao and Liu, Yiqun},
title = {LeCaRDv2: A Large-Scale Chinese Legal Case Retrieval Dataset},
year = {2024},
isbn = {9798400704314},
publisher = {Association for Computing Machinery},
address = {New York, NY, USA},
url = {https://doi.org/10.1145/3626772.3657887},
doi = {10.1145/3626772.3657887},
booktitle = {Proceedings of the 47th International ACM SIGIR Conference on Research and Development in Information Retrieval},
pages = {2251–2260},
numpages = {10},
location = {Washington DC, USA},
series = {SIGIR '24}
}

@inproceedings{li2025lexrag,
author = {Li, Haitao and Chen, Yifan and YiRan, Hu and Ai, Qingyao and Chen, Junjie and Yang, Xiaoyu and Yang, Jianhui and Wu, Yueyue and Liu, Zeyang and Liu, Yiqun},
title = {LexRAG: Benchmarking Retrieval-Augmented Generation in Multi-Turn Legal Consultation Conversation},
year = {2025},
isbn = {9798400715921},
publisher = {Association for Computing Machinery},
address = {New York, NY, USA},
url = {https://doi.org/10.1145/3726302.3730340},
doi = {10.1145/3726302.3730340},
booktitle = {Proceedings of the 48th International ACM SIGIR Conference on Research and Development in Information Retrieval},
pages = {3606–3615},
numpages = {10},
location = {Padua, Italy},
series = {SIGIR '25}
}

@misc{chen2024agentcourt,
      title={AgentCourt: Simulating Court with Adversarial Evolvable Lawyer Agents}, 
      author={Guhong Chen and Liyang Fan and Zihan Gong and Nan Xie and Zixuan Li and Ziqiang Liu and Chengming Li and Qiang Qu and Hamid Alinejad-Rokny and Shiwen Ni and Min Yang},
      year={2025},
      eprint={2408.08089},
      archivePrefix={arXiv},
      primaryClass={cs.CL},
      url={https://arxiv.org/abs/2408.08089}, 
}

@misc{chen2025agentmediation,
      title={Simulating Dispute Mediation with LLM-Based Agents for Legal Research}, 
      author={Junjie Chen and Haitao Li and Minghao Qin and Yujia Zhou and Yanxue Ren and Wuyue Wang and Yiqun Liu and Yueyue Wu and Qingyao Ai},
      year={2025},
      eprint={2509.06586},
      archivePrefix={arXiv},
      primaryClass={cs.CY},
      url={https://arxiv.org/abs/2509.06586}, 
}

@misc{moonshot2026kimik25,
      title={Kimi K2.5: Visual Agentic Intelligence}, 
      author={Kimi Team and Tongtong Bai and Yifan Bai and Yiping Bao and S. H. Cai and Yuan Cao and Y. Charles and H. S. Che and Cheng Chen and Guanduo Chen and Huarong Chen and Jia Chen and Jiahao Chen and Jianlong Chen and Jun Chen and Kefan Chen and Liang Chen and Ruijue Chen and Xinhao Chen and Yanru Chen and Yanxu Chen and Yicun Chen and Yimin Chen and Yingjiang Chen and Yuankun Chen and Yujie Chen and Yutian Chen and Zhirong Chen and Ziwei Chen and Dazhi Cheng and Minghan Chu and Jialei Cui and Jiaqi Deng and Muxi Diao and Hao Ding and Mengfan Dong and Mengnan Dong and Yuxin Dong and Yuhao Dong and Angang Du and Chenzhuang Du and Dikang Du and Lingxiao Du and Yulun Du and Yu Fan and Shengjun Fang and Qiulin Feng and Yichen Feng and Garimugai Fu and Kelin Fu and Hongcheng Gao and Tong Gao and Yuyao Ge and Shangyi Geng and Chengyang Gong and Xiaochen Gong and Zhuoma Gongque and Qizheng Gu and Xinran Gu and Yicheng Gu and Longyu Guan and Yuanying Guo and Xiaoru Hao and Weiran He and Wenyang He and Yunjia He and Chao Hong and Hao Hu and Jiaxi Hu and Yangyang Hu and Zhenxing Hu and Ke Huang and Ruiyuan Huang and Weixiao Huang and Zhiqi Huang and Tao Jiang and Zhejun Jiang and Xinyi Jin and Yu Jing and Guokun Lai and Aidi Li and C. Li and Cheng Li and Fang Li and Guanghe Li and Guanyu Li and Haitao Li and Haoyang Li and Jia Li and Jingwei Li and Junxiong Li and Lincan Li and Mo Li and Weihong Li and Wentao Li and Xinhang Li and Xinhao Li and Yang Li and Yanhao Li and Yiwei Li and Yuxiao Li and Zhaowei Li and Zheming Li and Weilong Liao and Jiawei Lin and Xiaohan Lin and Zhishan Lin and Zichao Lin and Cheng Liu and Chenyu Liu and Hongzhang Liu and Liang Liu and Shaowei Liu and Shudong Liu and Shuran Liu and Tianwei Liu and Tianyu Liu and Weizhou Liu and Xiangyan Liu and Yangyang Liu and Yanming Liu and Yibo Liu and Yuanxin Liu and Yue Liu and Zhengying Liu and Zhongnuo Liu and Enzhe Lu and Haoyu Lu and Zhiyuan Lu and Junyu Luo and Tongxu Luo and Yashuo Luo and Long Ma and Yingwei Ma and Shaoguang Mao and Yuan Mei and Xin Men and Fanqing Meng and Zhiyong Meng and Yibo Miao and Minqing Ni and Kun Ouyang and Siyuan Pan and Bo Pang and Yuchao Qian and Ruoyu Qin and Zeyu Qin and Jiezhong Qiu and Bowen Qu and Zeyu Shang and Youbo Shao and Tianxiao Shen and Zhennan Shen and Juanfeng Shi and Lidong Shi and Shengyuan Shi and Feifan Song and Pengwei Song and Tianhui Song and Xiaoxi Song and Hongjin Su and Jianlin Su and Zhaochen Su and Lin Sui and Jinsong Sun and Junyao Sun and Tongyu Sun and Flood Sung and Yunpeng Tai and Chuning Tang and Heyi Tang and Xiaojuan Tang and Zhengyang Tang and Jiawen Tao and Shiyuan Teng and Chaoran Tian and Pengfei Tian and Ao Wang and Bowen Wang and Chensi Wang and Chuang Wang and Congcong Wang and Dingkun Wang and Dinglu Wang and Dongliang Wang and Feng Wang and Hailong Wang and Haiming Wang and Hengzhi Wang and Huaqing Wang and Hui Wang and Jiahao Wang and Jinhong Wang and Jiuzheng Wang and Kaixin Wang and Linian Wang and Qibin Wang and Shengjie Wang and Shuyi Wang and Si Wang and Wei Wang and Xiaochen Wang and Xinyuan Wang and Yao Wang and Yejie Wang and Yipu Wang and Yiqin Wang and Yucheng Wang and Yuzhi Wang and Zhaoji Wang and Zhaowei Wang and Zhengtao Wang and Zhexu Wang and Zihan Wang and Zizhe Wang and Chu Wei and Ming Wei and Chuan Wen and Zichen Wen and Chengjie Wu and Haoning Wu and Junyan Wu and Rucong Wu and Wenhao Wu and Yuefeng Wu and Yuhao Wu and Yuxin Wu and Zijian Wu and Chenjun Xiao and Jin Xie and Xiaotong Xie and Yuchong Xie and Yifei Xin and Bowei Xing and Boyu Xu and Jianfan Xu and Jing Xu and Jinjing Xu and L. H. Xu and Lin Xu and Suting Xu and Weixin Xu and Xinbo Xu and Xinran Xu and Yangchuan Xu and Yichang Xu and Yuemeng Xu and Zelai Xu and Ziyao Xu and Junjie Yan and Yuzi Yan and Guangyao Yang and Hao Yang and Junwei Yang and Kai Yang and Ningyuan Yang and Ruihan Yang and Xiaofei Yang and Xinlong Yang and Ying Yang and Yi Yang and Yi Yang and Zhen Yang and Zhilin Yang and Zonghan Yang and Haotian Yao and Dan Ye and Wenjie Ye and Zhuorui Ye and Bohong Yin and Chengzhen Yu and Longhui Yu and Tao Yu and Tianxiang Yu and Enming Yuan and Mengjie Yuan and Xiaokun Yuan and Yang Yue and Weihao Zeng and Dunyuan Zha and Haobing Zhan and Dehao Zhang and Hao Zhang and Jin Zhang and Puqi Zhang and Qiao Zhang and Rui Zhang and Xiaobin Zhang and Y. Zhang and Yadong Zhang and Yangkun Zhang and Yichi Zhang and Yizhi Zhang and Yongting Zhang and Yu Zhang and Yushun Zhang and Yutao Zhang and Yutong Zhang and Zheng Zhang and Chenguang Zhao and Feifan Zhao and Jinxiang Zhao and Shuai Zhao and Xiangyu Zhao and Yikai Zhao and Zijia Zhao and Huabin Zheng and Ruihan Zheng and Shaojie Zheng and Tengyang Zheng and Junfeng Zhong and Longguang Zhong and Weiming Zhong and M. Zhou and Runjie Zhou and Xinyu Zhou and Zaida Zhou and Jinguo Zhu and Liya Zhu and Xinhao Zhu and Yuxuan Zhu and Zhen Zhu and Jingze Zhuang and Weiyu Zhuang and Ying Zou and Xinxing Zu},
      year={2026},
      eprint={2602.02276},
      archivePrefix={arXiv},
      primaryClass={cs.CL},
      url={https://arxiv.org/abs/2602.02276}, 
}

@misc{zeng2026glm5,
      title={GLM-5: from Vibe Coding to Agentic Engineering}, 
      author={GLM-5-Team and : and Aohan Zeng and Xin Lv and Zhenyu Hou and Zhengxiao Du and Qinkai Zheng and Bin Chen and Da Yin and Chendi Ge and Chenghua Huang and Chengxing Xie and Chenzheng Zhu and Congfeng Yin and Cunxiang Wang and Gengzheng Pan and Hao Zeng and Haoke Zhang and Haoran Wang and Huilong Chen and Jiajie Zhang and Jian Jiao and Jiaqi Guo and Jingsen Wang and Jingzhao Du and Jinzhu Wu and Kedong Wang and Lei Li and Lin Fan and Lucen Zhong and Mingdao Liu and Mingming Zhao and Pengfan Du and Qian Dong and Rui Lu and Shuang-Li and Shulin Cao and Song Liu and Ting Jiang and Xiaodong Chen and Xiaohan Zhang and Xuancheng Huang and Xuezhen Dong and Yabo Xu and Yao Wei and Yifan An and Yilin Niu and Yitong Zhu and Yuanhao Wen and Yukuo Cen and Yushi Bai and Zhongpei Qiao and Zihan Wang and Zikang Wang and Zilin Zhu and Ziqiang Liu and Zixuan Li and Bojie Wang and Bosi Wen and Can Huang and Changpeng Cai and Chao Yu and Chen Li and Chengwei Hu and Chenhui Zhang and Dan Zhang and Daoyan Lin and Dayong Yang and Di Wang and Ding Ai and Erle Zhu and Fangzhou Yi and Feiyu Chen and Guohong Wen and Hailong Sun and Haisha Zhao and Haiyi Hu and Hanchen Zhang and Hanrui Liu and Hanyu Zhang and Hao Peng and Hao Tai and Haobo Zhang and He Liu and Hongwei Wang and Hongxi Yan and Hongyu Ge and Huan Liu and Huanpeng Chu and Jia'ni Zhao and Jiachen Wang and Jiajing Zhao and Jiamin Ren and Jiapeng Wang and Jiaxin Zhang and Jiayi Gui and Jiayue Zhao and Jijie Li and Jing An and Jing Li and Jingwei Yuan and Jinhua Du and Jinxin Liu and Junkai Zhi and Junwen Duan and Kaiyue Zhou and Kangjian Wei and Ke Wang and Keyun Luo and Laiqiang Zhang and Leigang Sha and Liang Xu and Lindong Wu and Lintao Ding and Lu Chen and Minghao Li and Nianyi Lin and Pan Ta and Qiang Zou and Rongjun Song and Ruiqi Yang and Shangqing Tu and Shangtong Yang and Shaoxiang Wu and Shengyan Zhang and Shijie Li and Shuang Li and Shuyi Fan and Wei Qin and Wei Tian and Weining Zhang and Wenbo Yu and Wenjie Liang and Xiang Kuang and Xiangmeng Cheng and Xiangyang Li and Xiaoquan Yan and Xiaowei Hu and Xiaoying Ling and Xing Fan and Xingye Xia and Xinyuan Zhang and Xinze Zhang and Xirui Pan and Xu Zou and Xunkai Zhang and Yadi Liu and Yandong Wu and Yanfu Li and Yidong Wang and Yifan Zhu and Yijun Tan and Yilin Zhou and Yiming Pan and Ying Zhang and Yinpei Su and Yipeng Geng and Yong Yan and Yonglin Tan and Yuean Bi and Yuhan Shen and Yuhao Yang and Yujiang Li and Yunan Liu and Yunqing Wang and Yuntao Li and Yurong Wu and Yutao Zhang and Yuxi Duan and Yuxuan Zhang and Zezhen Liu and Zhengtao Jiang and Zhenhe Yan and Zheyu Zhang and Zhixiang Wei and Zhuo Chen and Zhuoer Feng and Zijun Yao and Ziwei Chai and Ziyuan Wang and Zuzhou Zhang and Bin Xu and Minlie Huang and Hongning Wang and Juanzi Li and Yuxiao Dong and Jie Tang},
      year={2026},
      eprint={2602.15763},
      archivePrefix={arXiv},
      primaryClass={cs.LG},
      url={https://arxiv.org/abs/2602.15763}, 
}

@misc{deepseek2025v32,
      title={DeepSeek-V3.2: Pushing the Frontier of Open Large Language Models}, 
      author={DeepSeek-AI and Aixin Liu and Aoxue Mei and Bangcai Lin and Bing Xue and Bingxuan Wang and Bingzheng Xu and Bochao Wu and Bowei Zhang and Chaofan Lin and Chen Dong and Chengda Lu and Chenggang Zhao and Chengqi Deng and Chenhao Xu and Chong Ruan and Damai Dai and Daya Guo and Dejian Yang and Deli Chen and Erhang Li and Fangqi Zhou and Fangyun Lin and Fucong Dai and Guangbo Hao and Guanting Chen and Guowei Li and H. Zhang and Hanwei Xu and Hao Li and Haofen Liang and Haoran Wei and Haowei Zhang and Haowen Luo and Haozhe Ji and Honghui Ding and Hongxuan Tang and Huanqi Cao and Huazuo Gao and Hui Qu and Hui Zeng and Jialiang Huang and Jiashi Li and Jiaxin Xu and Jiewen Hu and Jingchang Chen and Jingting Xiang and Jingyang Yuan and Jingyuan Cheng and Jinhua Zhu and Jun Ran and Junguang Jiang and Junjie Qiu and Junlong Li and Junxiao Song and Kai Dong and Kaige Gao and Kang Guan and Kexin Huang and Kexing Zhou and Kezhao Huang and Kuai Yu and Lean Wang and Lecong Zhang and Lei Wang and Liang Zhao and Liangsheng Yin and Lihua Guo and Lingxiao Luo and Linwang Ma and Litong Wang and Liyue Zhang and M. S. Di and M. Y Xu and Mingchuan Zhang and Minghua Zhang and Minghui Tang and Mingxu Zhou and Panpan Huang and Peixin Cong and Peiyi Wang and Qiancheng Wang and Qihao Zhu and Qingyang Li and Qinyu Chen and Qiushi Du and Ruiling Xu and Ruiqi Ge and Ruisong Zhang and Ruizhe Pan and Runji Wang and Runqiu Yin and Runxin Xu and Ruomeng Shen and Ruoyu Zhang and S. H. Liu and Shanghao Lu and Shangyan Zhou and Shanhuang Chen and Shaofei Cai and Shaoyuan Chen and Shengding Hu and Shengyu Liu and Shiqiang Hu and Shirong Ma and Shiyu Wang and Shuiping Yu and Shunfeng Zhou and Shuting Pan and Songyang Zhou and Tao Ni and Tao Yun and Tian Pei and Tian Ye and Tianyuan Yue and Wangding Zeng and Wen Liu and Wenfeng Liang and Wenjie Pang and Wenjing Luo and Wenjun Gao and Wentao Zhang and Xi Gao and Xiangwen Wang and Xiao Bi and Xiaodong Liu and Xiaohan Wang and Xiaokang Chen and Xiaokang Zhang and Xiaotao Nie and Xin Cheng and Xin Liu and Xin Xie and Xingchao Liu and Xingkai Yu and Xingyou Li and Xinyu Yang and Xinyuan Li and Xu Chen and Xuecheng Su and Xuehai Pan and Xuheng Lin and Xuwei Fu and Y. Q. Wang and Yang Zhang and Yanhong Xu and Yanru Ma and Yao Li and Yao Li and Yao Zhao and Yaofeng Sun and Yaohui Wang and Yi Qian and Yi Yu and Yichao Zhang and Yifan Ding and Yifan Shi and Yiliang Xiong and Ying He and Ying Zhou and Yinmin Zhong and Yishi Piao and Yisong Wang and Yixiao Chen and Yixuan Tan and Yixuan Wei and Yiyang Ma and Yiyuan Liu and Yonglun Yang and Yongqiang Guo and Yongtong Wu and Yu Wu and Yuan Cheng and Yuan Ou and Yuanfan Xu and Yuduan Wang and Yue Gong and Yuhan Wu and Yuheng Zou and Yukun Li and Yunfan Xiong and Yuxiang Luo and Yuxiang You and Yuxuan Liu and Yuyang Zhou and Z. F. Wu and Z. Z. Ren and Zehua Zhao and Zehui Ren and Zhangli Sha and Zhe Fu and Zhean Xu and Zhenda Xie and Zhengyan Zhang and Zhewen Hao and Zhibin Gou and Zhicheng Ma and Zhigang Yan and Zhihong Shao and Zhixian Huang and Zhiyu Wu and Zhuoshu Li and Zhuping Zhang and Zian Xu and Zihao Wang and Zihui Gu and Zijia Zhu and Zilin Li and Zipeng Zhang and Ziwei Xie and Ziyi Gao and Zizheng Pan and Zongqing Yao and Bei Feng and Hui Li and J. L. Cai and Jiaqi Ni and Lei Xu and Meng Li and Ning Tian and R. J. Chen and R. L. Jin and S. S. Li and Shuang Zhou and Tianyu Sun and X. Q. Li and Xiangyue Jin and Xiaojin Shen and Xiaosha Chen and Xinnan Song and Xinyi Zhou and Y. X. Zhu and Yanping Huang and Yaohui Li and Yi Zheng and Yuchen Zhu and Yunxian Ma and Zhen Huang and Zhipeng Xu and Zhongyu Zhang and Dongjie Ji and Jian Liang and Jianzhong Guo and Jin Chen and Leyi Xia and Miaojun Wang and Mingming Li and Peng Zhang and Ruyi Chen and Shangmian Sun and Shaoqing Wu and Shengfeng Ye and T. Wang and W. L. Xiao and Wei An and Xianzu Wang and Xiaowen Sun and Xiaoxiang Wang and Ying Tang and Yukun Zha and Zekai Zhang and Zhe Ju and Zhen Zhang and Zihua Qu},
      year={2025},
      eprint={2512.02556},
      archivePrefix={arXiv},
      primaryClass={cs.CL},
      url={https://arxiv.org/abs/2512.02556}, 
}

@misc{yang2025qwen3,
      title={Qwen3 Technical Report}, 
      author={An Yang and Anfeng Li and Baosong Yang and Beichen Zhang and Binyuan Hui and Bo Zheng and Bowen Yu and Chang Gao and Chengen Huang and Chenxu Lv and Chujie Zheng and Dayiheng Liu and Fan Zhou and Fei Huang and Feng Hu and Hao Ge and Haoran Wei and Huan Lin and Jialong Tang and Jian Yang and Jianhong Tu and Jianwei Zhang and Jianxin Yang and Jiaxi Yang and Jing Zhou and Jingren Zhou and Junyang Lin and Kai Dang and Keqin Bao and Kexin Yang and Le Yu and Lianghao Deng and Mei Li and Mingfeng Xue and Mingze Li and Pei Zhang and Peng Wang and Qin Zhu and Rui Men and Ruize Gao and Shixuan Liu and Shuang Luo and Tianhao Li and Tianyi Tang and Wenbiao Yin and Xingzhang Ren and Xinyu Wang and Xinyu Zhang and Xuancheng Ren and Yang Fan and Yang Su and Yichang Zhang and Yinger Zhang and Yu Wan and Yuqiong Liu and Zekun Wang and Zeyu Cui and Zhenru Zhang and Zhipeng Zhou and Zihan Qiu},
      year={2025},
      eprint={2505.09388},
      archivePrefix={arXiv},
      primaryClass={cs.CL},
      url={https://arxiv.org/abs/2505.09388}, 
}

@misc{qwen2026qwen35,
  title        = {Qwen3.5: Towards Native Multimodal Agents},
  author       = {{Qwen Team}},
  year         = {2026},
  howpublished = {\url{https://qwen.ai/blog?id=qwen3.5}}
}

@incollection{Brenner_Whitmeyer_2009,
  author    = {Brenner, Saul and Whitmeyer, Joseph M.},
  title     = {The Legal Model},
  booktitle = {Strategy on the United States Supreme Court},
  publisher = {Cambridge University Press},
  address   = {Cambridge},
  year      = {2009},
  pages     = {3--10}
}

@article{Rachlinski2015NumericJudgments,
  author  = {Rachlinski, Jeffrey J. and Wistrich, Andrew J. and Guthrie, Chris},
  title   = {Can Judges Make Reliable Numeric Judgments? Distorted Damages and Skewed Sentences},
  journal = {Indiana Law Journal},
  volume  = {90},
  number  = {2},
  pages   = {695--739},
  year    = {2015},
  url     = {https://www.repository.law.indiana.edu/ilj/vol90/iss2/6}
}

@article{ChangChenLiaoLin2023AskMoreAwardedMore,
  author  = {Chang, Yun-Chien and Chen, Kong-Pin and Liao, Jen-Che and Lin, Chang-Ching},
  title   = {Ask More, Awarded More: Evidence from Taiwan's Courts},
  journal = {International Review of Law and Economics},
  volume  = {76},
  pages   = {106171},
  year    = {2023},
  doi     = {10.1016/j.irle.2023.106171},
  url = {https://www.sciencedirect.com/science/article/pii/S0144818823000492}
}

@article{WojciechowskiWhiteAllefeldPothos2025OrderEffects,
  author  = {Wojciechowski, Bartosz W. and White, Lee C. and Allefeld, Carsten and Pothos, Emmanuel M.},
  title   = {Order Effects and the Evaluation Bias in Legal Decision Making},
  journal = {Decision},
  volume  = {12},
  number  = {3},
  pages   = {246--267},
  year    = {2025},
  doi     = {10.1037/dec0000263}
}

@incollection{Fay2024OutOfBounds,
  author    = {Fay, Sydney A.},
  title     = {{``Out of Bounds'': The Influence of Personal and Institutionalized Bounded Rationality on Judicial Decision Making}},
  booktitle = {Using Organizational Theory to Study, Explain, and Understand Criminal Legal Organizations},
  publisher = {Springer Nature Switzerland},
  year      = {2024},
  pages     = {17--33},
  doi       = {10.1007/978-3-031-66285-0_2}
}

@article{ColauxSchiffinoMoyson2023Managerialization,
  author  = {Colaux, {\'E}milien and Schiffino, Nathalie and Moyson, St{\'e}phane},
  title   = {Neither the Magic Bullet Nor the Big Bad Wolf: A Systematic Review of Frontline Judges' Attitudes and Coping Regarding Managerialization},
  journal = {Administration \& Society},
  volume  = {55},
  number  = {5},
  pages   = {921--952},
  year    = {2023},
  doi     = {10.1177/00953997231157748}
}

@article{BlackOwensWedekingWohlfarth2016OpinionClarity,
  author  = {Black, Ryan C. and Owens, Ryan J. and Wedeking, Justin and Wohlfarth, Patrick C.},
  title   = {The Influence of Public Sentiment on Supreme Court Opinion Clarity},
  journal = {Law \& Society Review},
  volume  = {50},
  number  = {3},
  pages   = {703--732},
  year    = {2016}
}




% LNCS Homepage, \url{http://www.springer.com/lncs}, last accessed 2023/10/25
% \end{thebibliography}
\end{document}